\documentclass[a4paper,fleqn]{cas-sc}
\usepackage[numbers]{natbib}

\usepackage[utf8]{inputenc}
\usepackage{amsmath}
\usepackage{amsfonts}
\usepackage{amssymb}
\usepackage{amsthm}
\usepackage{float}
\usepackage{caption}
\usepackage{subcaption}
\usepackage[algosection,linesnumbered,ruled,lined,boxed]{algorithm2e}
\SetArgSty{}

\SetCommentSty{mycommfont}

\newcommand\bsf{best-so-far}
\newcommand\cmax{Cmax}

\begin{document}
\let\WriteBookmarks\relax
\def\floatpagepagefraction{1}
\def\textpagefraction{.001}
\shorttitle{Iterative beam search algorithms for the permutation flowshop}
\shortauthors{L. Libralesso et al.}

\title [mode = title]{Iterative beam search algorithms for the permutation flowshop}         

\author[1]{Luc Libralesso}
\ead{luc.libralesso@grenoble-inp.fr}
\cormark[1]

\author[1]{Pablo Andres Focke}
\ead{pablofocke@gmail.com}

\author[1]{Aurélien Secardin}
\ead{aurelien.secardin@icloud.com}

\author[1]{Vincent Jost}
\ead{vincent.jost@grenoble-inp.fr}

\address[1]{Univ. Grenoble Alpes, CNRS, Grenoble INP, G-SCOP, 38000 Grenoble, France}

\begin{abstract}
We study an iterative beam search algorithm for the permutation flowshop (makespan and flowtime minimization). This algorithm combines branching strategies inspired by recent branch-and-bounds and a guidance strategy inspired by the LR heuristic. It obtains competitive results, reports many new-best-so-far solutions on the VFR benchmark (makespan minimization) and the Taillard benchmark (flowtime minimization) without using any NEH-based branching or iterative-greedy strategy. \\The source code is available at: \url{https://gitlab.com/librallu/cats-pfsp}.
\end{abstract}

\begin{highlights}

\item First use of an iterative beam search for the permutation flowshop

\item Simple yet efficient heuristic guidance strategies

\item Bi-directional search strategy to minimize the makespan variant

\item New \bsf\ solutions on VRF instances (makespan, 105/160 open instances)

\item New \bsf\ solutions on Taillard instances (flowtime, 55/100 open instances)

\end{highlights}


\maketitle

\section{Introduction}

In the flowshop problem, one has to schedule jobs, where each job has to follow the same route of machines. The goal is to find a job order that minimizes some criteria. The permutation flowshop, also called PFSP, is a common (and fundamental) variant that imposes the machines to process jobs in the same order (thus, a permutation of jobs is enough to describe a solution). The permutation flowshop has been one of the most studied problems in the literature \cite{reza2005flowshop,pan2013comprehensive} and has been considered on various industrial applications \cite{krajewski1987kanban,vakharia1990designing}. We may also note that the permutation flowshop is at the origin of multiple other variants, for instance, the blocking permutation flowshop \cite{wang2011hybrid}, the multiobjective permutation flowshop \cite{li2008effective}, the distributed permutation flowshop \cite{gao2011hybrid}, the no-idle permutation flowshop \cite{pan2014effective}, the permutation flowshop with buffers \cite{nowicki1999permutation} and many others. Regarding the criteria to minimize, we study in this paper, two of the most studied objectives: the makespan (minimizing the completion time of the last job on the last machine) and the flowtime (minimizing the sum of completion times of each job on the last machine). According to the scheduling notation introduced by Graham, Lawler, Lenstra, and Rinnooy Kan \cite{graham1979optimization}, the makespan criterion is denoted $F_m|prmu|\cmax$ and the flowtime criterion $F_m|prmu|\sum C_i$.

\paragraph{}
Consider the following example instance with $m=3$ machines with $n=4$ jobs ($j_1,j_2,j_3,j_4$) with the job processing time matrix $P$ defined as follows where $P_{j,m}$ indicates the processing time of job $j$ on machine $m$:

\[
    P = \begin{pmatrix}
        3 & 2 & 1 & 3 \\
        3 & 4 & 3 & 1 \\
        2 & 1 & 3 & 2 \\
    \end{pmatrix}
\]

\begin{figure}[pos=ht]
    \centering
    \includegraphics{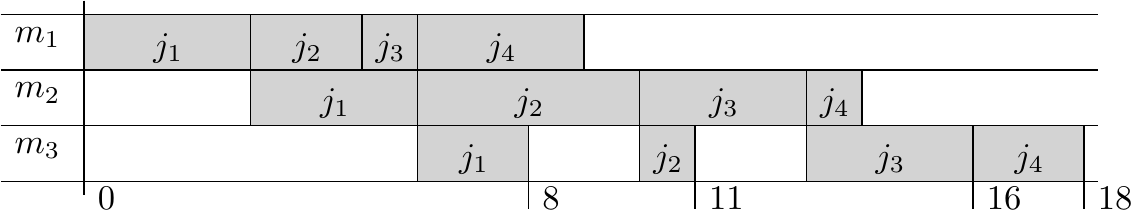}
    \caption{A solution for the example instance with a job order $\sigma = j_1, j_2, j_3, j_4$}
    \label{fig:exsol}
\end{figure}

One possible solution can be described in Figure \ref{fig:exsol}. This solution has a makespan (completion time of the last job on the last machine) of 18 and a flowtime (sum of completion times on the last machine) of $8+11+16+18=53$.

\paragraph{} Regarding resolution methods, the makespan minimization permutation flowshop problem has been massively studied over the last 50 years and a large number of numerical methods have been applied. 

In 1983, Nawaz, Enscore, Ham proposed an insertion based heuristic (later called NEH) \cite{nawaz1983heuristic}. This heuristic sorts jobs by some criterion (usually by a non-decreasing sum of processing times), then adds them one by one at the position that minimizes the objective function. The NEH, obtained, at the time, excellent results compared to other heuristics and can be used to perform greedy algorithms and perturbation-based algorithms as well. It has been largely considered as an essential component in order to solve large-scale permutation flowshop instances, and multiple methods have been built using it. One of the most famous ones is the Taillard's acceleration \cite{taillard1990some}, that reduces the cost of inserting a job at all possible positions from $O(n^2.k)$ to $O(n.k)$. Considering these results, multiple works aim to improve the NEH heuristic \cite{framinan2003efficient,nagano2002high,kalczynski2008improved,dong2008improved,ribas2009improvement,vasiljevic2015handling,liu2017new} to cite a few.

The (meta-)heuristics state-of-the-art methods for the makespan minimization usually perform an iterated-greedy algorithm \cite{ruiz2007simple,fernandez2019best}. Such algorithms start with a NEH heuristic to build an initial solution. Then, destroy a part of it and reconstruct it using again an NEH heuristic. To the best of our knowledge, the current state-of-the-art for the makespan minimization criterion is the variable block insertion heuristic \cite{kizilay_variable_2019}. The variable block insertion heuristic starts by finding an initial solution using the FRB5 heuristic \cite{rad2009new}. It removes some block of jobs, applies a local search procedure, then reinserts the block in the best possible position. We may note that other algorithms exist to solve the makespan minimization. To cite a few, we can find some hybrid algorithms \cite{zheng2003effective} (a combination of the NEH heuristic as a part of the initial population, a genetic algorithm, and simulated annealing to replace the mutation), memetic algorithms \cite{kurdi2020memetic}, an automatically designed local-search scheme \cite{pagnozzi2019automatic}.

The (meta-)heuristics methods for the flowtime minimization also involve the NEH heuristic, but also some other constructive methods as well. For instance, the Liu and Reeve's method (LR) \cite{liu2001constructive}. This method performs a forward search (\emph{i.e} appending jobs at the end of the partial schedule). It was later improved to reduce its complexity from $O(n^3m)$ to $O(n^2m)$, later called the FF algorithm \cite{fernandez2015new}. Later, this scheme was integrated into a beam search algorithm (more on that later) that obtained state-of-the-art performance \cite{fernandez2017beam}. Recently, this beam search was integrated within a biased random-key genetic algorithm as a warm-start procedure \cite{andrade2019minimizing}. This is, to the best of our knowledge, the state-of-the-art method for the flowtime minimization.

\paragraph{}
Regarding exact-methods, a recent branch \& bound \cite{gmys2020computationally} brought light on a bi-directional  branching (\emph{i.e} constructing the candidate solution from the beginning and the end at the same time) combined with a simple yet efficient bounding scheme to solve the makespan minimization criterion. The resulting branch \& bound obtained excellent performance and was even able to solve to optimality almost all large VFR instances with 20 machines.

Moreover, recently, an iterative beam search has been proposed and, successfully applied to various combinatorial optimization problems as guillotine 2D packing problems \cite{libralesso2020anytime,fontan2020packingsolver}, the sequential ordering problem \cite{libralesso2019tree} and the longest common subsequence problem \cite{libralesso:hal-02895115}. This iterative beam search scheme, at the beginning of the search, behaves as a greedy algorithm and more and more as a branch \& bound algorithm as time goes (it performs a series of beam search iterations with a geometric growth). It naturally combines search-space reductions from branch \& bounds and guidance strategies from classical (meta-)heuristics. Considering the success of recent branch \& bound branching schemes and the performance of greedy-like algorithms to solve the permutation flowshop, it would be a natural idea to combine them. However, to the best of our knowledge, it has not been studied before. This paper aims to fill this gap. For the makespan criterion, we implemented a bi-directional branching scheme and combined it with a variant of the LR \cite{liu2001constructive} guidance strategy and use an iterative beam-search algorithm to perform the search. We report competitive results and find new best-known solutions on many large VFR instances (we improve the best-known solution for almost all instances with 500 jobs or more and 40 machines or more). Note that these results are interesting and new as almost all the efficient algorithms in the literature are based on the NEH heuristic or the iterated greedy algorithm. This is not the case for our algorithm as it is based on a variant of the LR heuristic and an exact-method branching scheme (bi-directional branching). 

Regarding the flowtime criterion, the bi-directional branching cannot be directly applied (the bounding procedure is less efficient than for the makespan criterion). However, we show that an iterative beam search with a simple forward search (modified LR algorithm) is efficient, and, reports new best-solutions for the Taillard's benchmark (almost all solutions for instances with 100 jobs or more were improved).

This paper is structured as follows: Section \ref{sec:branching} presents the branching schemes we implement (the forward and bi-directional search) for both criteria (makespan and flowtime). Section \ref{sec:guides} present the guides we implement (the bound guide, the idle-time guide and mixes between these two first guides). Section \ref{sec:search} presents the iterative beam search strategy and Section \ref{sec:num} presents the results obtained by running all variants described in this paper, showing that an iterative beam search combined with a simple variant of the LR heuristic can outperform the state-of-the-art.

\section{Branching schemes}\label{sec:branching}

We present in this section the two branching schemes we use (\emph{i.e.} the search tree structure): the forward search (\emph{i.e} constructing the solution from the beginning) and the bi-directional search (\emph{i.e.} constructing the solution from the beginning and the end).

\subsection{Forward branching}

The forward branching assigns jobs at the first free position in the partial sequences (it constructs the solutions from the beginning). The root corresponds to a situation where the candidate solution contains no job (\emph{i.e.} $c.\text{\textsc{starting}} = \emptyset$). Each of the search-tree node corresponds to the first jobs in the resulting solution. Children of a given node correspond to a possible insertion of each job that is not scheduled yet at the end of the schedule. Each node stores information about the partial candidate solution (jobs already added), the release time of each machine, and the partial makespan (resp. flowtime). A candidate solution (or node) $c$ is considered as ``goal'' or ``feasible'' if all jobs are inserted (\emph{i.e.} $c.\text{\textsc{starting}} = J$) and contains the following information:
\begin{itemize}
    \item \textsc{starting}: vector of jobs inserted that lead to the candidate $c$ (fist jobs of the sequence we want to generate).
    \item \textsc{frontStarting}: vector of times where machines are first available after appending \textsc{starting} jobs.
\end{itemize}

Before presenting the forward children-generation, we present how to insert a job $j \in J$ in a candidate solution $c$ (Algorithm \ref{alg:forwardinsert}). This insertion can be done in $O(m)$ where $m$ is the number of machines.

\begin{algorithm}[ht]
    \SetAlgoLined
    \DontPrintSemicolon
    \SetNoFillComment
	\SetKwInOut{Input}{Input}
    \Input{candidate solution (or node) $c$}
    \Input{job to be inserted $j \in J$}
    \BlankLine
    $c.\text{\textsc{frontStarting}}_1 \gets c.\text{\textsc{frontStarting}}_1 + P_{j,1}$\;
    \For{$i \in \{2, \dots m\}$}{
        \eIf{$c.\text{\textsc{frontStarting}}_{i-1} > c.\text{\textsc{frontStarting}}_{i}$}{
            \tcc{there is some idle time on machine $i$}
            $\text{idle} \gets c.\text{\textsc{frontStarting}}_{i-1} - c.\text{\textsc{frontStarting}}_{i}$\;
            $c.\text{\textsc{frontStarting}}_{i} \gets c.\text{\textsc{frontStarting}}_{i-1} + P_{j,i}$\;
        }{
            \tcc{no idle time on machine $i$}
            $c.\text{\textsc{frontStarting}}_{i} \gets c.\text{\textsc{frontStarting}}_{i} + P_{j,i}$\;
        }
    }
    $c.\text{\textsc{starting}} \gets c.\text{\textsc{starting}} \cup \{j\}$\;
    \caption{Forward search: insertion of job $j$ in candidate solution $c$ (\textsc{InsertForward}$(c,j)$)\label{alg:forwardinsert}}
\end{algorithm}

Algorithm \ref{alg:forward} presents the forward branching pseudo-code (how to generate all children of a candidate solution $c$). 

\begin{algorithm}[ht]
    \SetAlgoLined
    \DontPrintSemicolon
    \SetNoFillComment
	\SetKwInOut{Input}{Input}
    \Input{candidate solution (or node) $c$}
    \BlankLine
    $\text{children} \gets \emptyset$\;
    \For{$j \in \text{unscheduled jobs}$}{
        $\text{children} \gets \text{children} \cup \textsc{InsertForward}(\text{Copy}(c), j)$\;
    }
    \Return{children}\;
    \caption{Forward search children generation from a candidate solution $c$ (\textsc{Children}$(c)$) \label{alg:forward}}
\end{algorithm}

\subsection{Bi-directional branching}

To the best of our knowledge, the bi-directional branching was first introduced in 1980 \cite{potts1980adaptive}. The bi-directional search appends jobs at the beginning and the end of the candidate solution. It aims to exploit the property of the inverse problem (job order inversed and machine order inversed). Since then, the efficiency of this scheme has been largely recognized to solve the makespan minimization optimally \cite{carlier1996two,ladhari2005computational,lemesre2007exact,drozdowski2011grid,chakroun2013combining,ritt2016branch}. Recently, a parallel branch \& bound was successfully used to solve the makespan minimization criterion \cite{gmys2020computationally} using this bi-directional scheme. Multiple ways to decide if the algorithm performs a forward or backward insertion were studied (for instance alternating between a forward insertion and backward insertion). This study found out that the best way is selecting the insertion type that has the less remaining children after the bounding pruning step. Ties are broken by selecting the type of insertion that maximizes the sum of the lower bounds as large lower bounds are usually a more precise estimation.

A candidate solution (or node) $c$ is considered as ``goal'' or ``feasible'' if all jobs are inserted (\emph{i.e.} $c.\textsc{starting} \cup c.\textsc{finishing} = J$) and contains the following information:
\begin{itemize}
    \item \textsc{starting}: vector of jobs inserted at the beginning of the partial permutation that lead to the candidate $c$ (first jobs of the sequence we want to generate).
    \item \textsc{frontStarting}: vector of times where machines are first available after appending \textsc{starting} jobs.
    \item \textsc{finishing}: (inverted) vector of jobs inserted at the end of the partial permutation that lead to the candidate $c$ (last jobs of the sequence we want to generate).
    \item \textsc{frontFinishing}: vector of times where machines are no more available after appending \textsc{starting} jobs.
\end{itemize}

Algorithm \ref{alg:bidirectional} presents the bi-directional branching pseudo-code. We use \textsc{insertForward} (Algorithm \ref{alg:forwardinsert}) to insert a job within the \textsc{starting} vector and \textsc{insertBackward} that inserts a job within the \textsc{finishing} vector. This procedure is almost similar to \textsc{insertForward} but iterates over machines in an inverted order ($m \rightarrow 2$ instead of $2 \rightarrow m$). It generates children of both the forward and backward search (lines 1-6), prunes nodes that are dominated by the best-known solution (or upper-bound, lines 7-8). Then, it chooses the scheme that has fewer children (thus, usually a smaller search-space) and breaks ties by selecting the scheme having the more precise lower bounds (sum of lower bounds).

\begin{algorithm}[ht]
    \SetAlgoLined
    \DontPrintSemicolon
    \SetNoFillComment
	\SetKwInOut{Input}{Input}
    \Input{candidate solution (or node) $c$}
    \BlankLine
    $\text{F} \gets \emptyset$ \tcc*{F correspond to the children obtained by forward search}
    $\text{B} \gets \emptyset$ \tcc*{B correspond to the children obtained by backward search}
    \For{$j \in \text{unscheduled jobs}$}{
        $\text{F} \gets \text{F} \cup \textsc{InsertForward}(\text{Copy}(c), j)$\;
        $\text{B} \gets \text{B} \cup \textsc{InsertBackward}(\text{Copy}(c), j)$\;
    }
    $\text{F} \gets \{ c | c \in \text{F} \;\;\text{if}\;\; \textsc{Bound}(c) < \text{best known solution} \}$ \tcc*{removing forward nodes dominated by the UB}
    $\text{B} \gets \{ c | c \in \text{B} \;\;\text{if}\;\; \textsc{Bound}(c) < \text{best known solution} \}$ \tcc*{removing backward nodes dominated by the UB}
    \eIf{$|\text{F}| < |\text{B}| \;\;\;\lor\;\;\; (|\text{F}| = |\text{B}| \;\;\land\;\; \sum_{c \in F} \textsc{Bound}(c) > \sum_{c \in B} \textsc{Bound}(c)) $}{
        \Return{F} \tcc{chosing the forward search}
    }{
        \Return{B} \tcc{chosing the backward search}
    }
    \caption{Bi-directional search children generation from a candidate solution $c$ (\textsc{Children}$(c)$) \label{alg:bidirectional}}
\end{algorithm}

\section{Guides}\label{sec:guides}

In the previous section, we discussed the branching rules that define a search tree. As such trees are usually large, a way to tell which node is apriori more desirable is needed. In branch-and-bounds, this mechanism is called ``bound'' and also constitutes an optimistic estimate of the best solution that can be achieved in a given sub-tree. In constructive meta-heuristics, the guidance strategy is usually not an optimistic estimate which often allows finding better solutions (for instance the LR \cite{liu2001constructive} greedy guidance strategy). In this section, we present several guidance strategies for both the makespan and flowtime criteria.

\subsection{Bound}

We define the bound guidance strategy for the forward search and makespan minimization as follows. It measures the first time the last machine (machine $m$) is available and assumes that each remaining job can be scheduled without any idle time. 

\[ \text{F\ } g_1 = \cmax_{f,m} + R_m \]

The bound guidance strategy for the bi-directional search and makespan minimization is defined as follows. It generalizes the bound for the forward search by also taking into account the backward front. We may note that the bi-directional branching allows computing a better bound as all machines are relevant for this bound (compared to the forward branching bound in which only the last machine is used to compute a bound).

\[ \text{FB\ } g_1 = \max_{i \in M} ( \cmax_{f,i} + R_i + \cmax_{b,i} )\]

The flowtime bound is defined as the sum of end times for each job scheduled in the forward search. Each time a job is added to the candidate solution, the flowtime value is modified.


\subsection{idle time}

The bound guide is an effective guidance strategy, but is known to be imprecise at the beginning of the search (\emph{i.e.} the first levels of the search tree). Another guide that is usually considered as a part of effective greedy strategies (for instance the LR heuristic) is to use the idle time of the partial solution. Usually, a solution with a small idle time reaches good performance on both the makespan or flowtime criteria.

The idle time can be defined as follows:

\[ \text{FB\ } g_2 = \sum_{i \in M} I_{f,i} + I_{b,i} \]

\subsection{bound and idle time}

As it is noted in many works \cite{liu2001constructive,fernandez2017beam}, another interesting guidance strategy is to combine both guidance strategies discussed earlier (\emph{i.e.} the bound and idle time guides). Indeed, while the bound guide is usually ineffective to guide the search close to the root, it is very precise close to feasible solutions. Inversely, the idle time is an efficient guide close to the root but relatively inefficient close to feasible solutions. We study the \emph{bound and idle time guide} that linearly reduces the contribution of the idle time to favor the bound depending on the completion level of the candidate solution.

The bound and idle time guide can be defined as follows, where $C$ is a value used to make the idle time and bound comparable:

\[ g_3 =  \alpha \;.\; g_1 + (1-\alpha) \;.\; C \;.\; g_2 \]

where $\alpha$ corresponds to the proportion of jobs added (\emph{i.e.} 0 if no jobs are added, 1 if all jobs are added). It is defined as follows: $\alpha = \frac{|F|+|B|}{|J|}$ for the bi-directional branching or $\alpha = \frac{|F|}{|J|}$ for the forward branching.

\subsection{bound and weighted idle time}

Another useful remark found in greedy algorithms for the permutation flowshop problem \cite{liu2001constructive} is to add additional weight to the idle time produced by the first machines at the beginning of the search (as it will have a greater impact on the objective function than the others). However, the LR heuristic cannot be directly applied in a general tree search context. Indeed, it is sometimes noted \cite{fernandez2017beam} that algorithms like the beam search usually compare nodes from different parents, thus, it is needed to adapt the LR heuristic guidance that only compares nodes with the same parent. We propose two different simple yet efficient ways to implement similar ideas. The search is guided by a combination of a weighted idle time and by the bounding procedure.

The first guide, used for the forward search for the flowtime minimization is defined as follows, where $I_w$ is the weighted idle time and $C = m . \frac{\sum_{i \in M} I_i}{2}$ :

\[ \text{F\ } g_4 = \alpha \;.\; g_1 + (1-\alpha) \;.\; ( I_w + C) \]

At each time we add a job $j$ to the end of the partial solution, we increase the weighted idle times as follows where $v$ is the idle time added by the job $j$ in machine $i$:

\[ I_w = I_w + v \;.\; (\alpha\;.\;(m-i) + 1) \]

For the bi-directional branching, we present a new guidance strategy that considers the sum of idle time percentage for each front. Doing this, it allows making idle time on the first machines more important to the forward search and the idle time on the last machines more important to the backward search. The bound and weighted idle time guide for the bi-directional search is defined as follows:

\[ \text{FB\ } g_4 =  (1-\alpha) . g_1 . \left( \sum_{i \in M} \frac{I_{f,i}}{\cmax_{f,i}} +\frac{I_{b,i}}{\cmax_{b,i}} \right)  + \alpha . g_1 \]













\section{The search strategy: Iterative beam search}\label{sec:search}

\begin{figure}[pos=ht]
    \centering
    \begin{subfigure}[b]{0.30\textwidth}
        \centering \includegraphics[width=\textwidth]{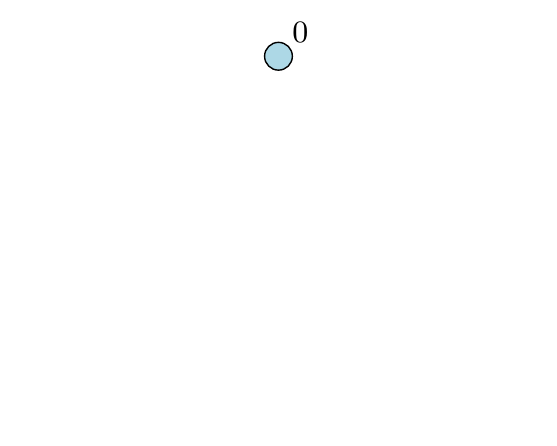}
    \end{subfigure}
    ~
    \begin{subfigure}[b]{0.30\textwidth}
        \centering \includegraphics[width=\textwidth]{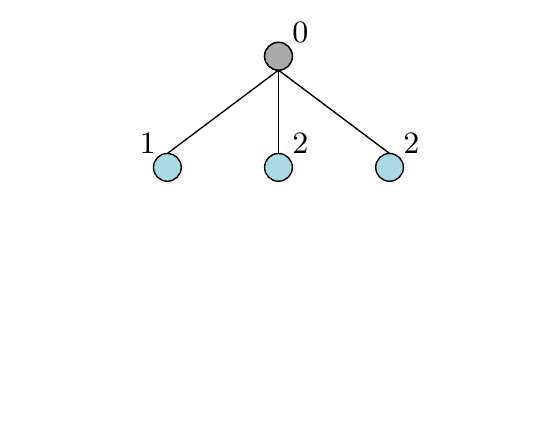}
    \end{subfigure}
    ~
    \begin{subfigure}[b]{0.30\textwidth}
        \centering \includegraphics[width=\textwidth]{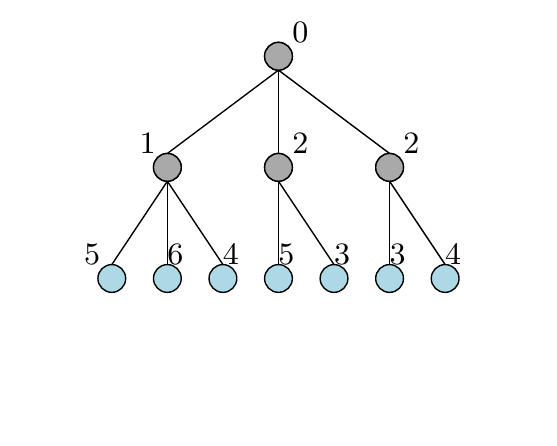}
    \end{subfigure}

    \begin{subfigure}[b]{0.30\textwidth}
        \centering \includegraphics[width=\textwidth]{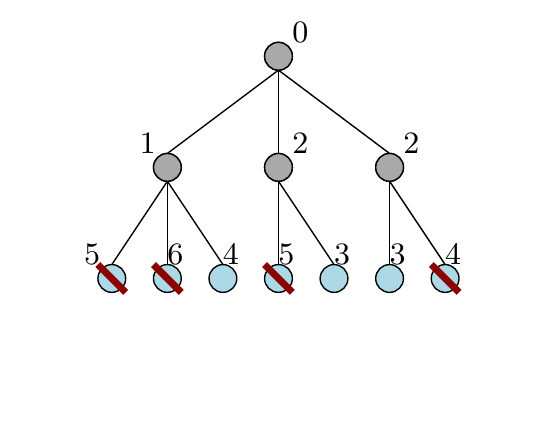}
    \end{subfigure}
    ~
    \begin{subfigure}[b]{0.30\textwidth}
        \centering \includegraphics[width=\textwidth]{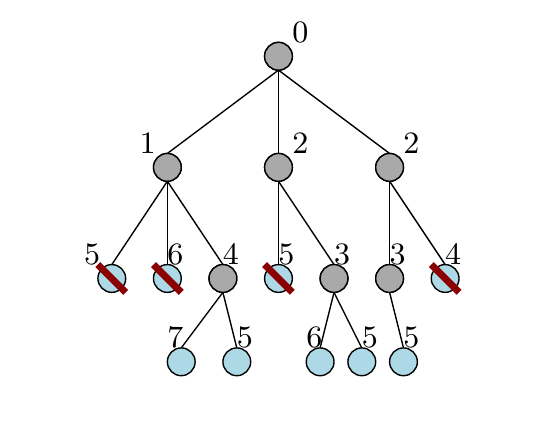}
    \end{subfigure}
    ~
    \begin{subfigure}[b]{0.30\textwidth}
        \centering \includegraphics[width=\textwidth]{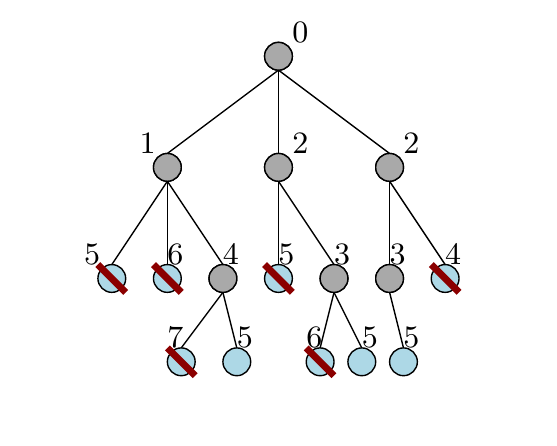}
    \end{subfigure}
    \caption{Beam Search Iterations with a beam width $D=3$ }\label{fig:bs_iterations}
\end{figure}

Beam Search is a tree search algorithm that uses a parameter called the beam size ($D$). Beam Search behaves like a truncated \emph{Breadth First Search (BrFS)}. It only considers the best $D$ nodes on a given level. The other nodes are discarded. Usually, we use the bound of a node to choose the most promising nodes. It generalizes both a greedy algorithm (if $D=1$) and a BrFS (if $D=\infty$). Figure \ref{fig:bs_iterations} presents an example of beam search execution with a beam width $D=3$.

\paragraph{}
\emph{Beam Search} was originally proposed in \cite{reddy1977speech} and used in speech recognition. It is an incomplete (\emph{i.e.} performing a partial tree exploration and can miss optimal solutions) tree search parametrized by the beam width $D$. Thus, it is not an anytime algorithm. The parameter $D$ allows controlling the quality of the solutions and the execution time. The larger $D$ is, the longer it will take to reach feasible solutions, and the better these solutions will be.

\paragraph{}
Recently, a variant of beam search, called iterative beam search, was proposed and obtained state-of-the-art results on various combinatorial optimization problems \cite{libralesso2019tree,libralesso2020anytime,libralesso:hal-02895115,fontan2020packingsolver}. Iterative beam search performs a series of restarting beam search with geometrically increasing beam size until the time limit is reached.
Algorithm \ref{alg:BS} shows the pseudo-code of an iterative beam search. The algorithm runs multiple beam searches starting with $D=1$ (line 1) and increases the beam size (line 8) geometrically. Each run explores the tree with the given parameter $D$. In the pseudo-code, we increase geometrically the beam size by 2. This parameter can be tuned, however, we did not notice a significant variation in the performance while adjusting this parameter. This parameter (that can be a real number) should be strictly larger than 1 (for the beam to expand) and should not be too large, say less than 3 or 5 (otherwise, the beam grows too fast and when the time limit is reached, most of the computational time was possibly wasted in the last incomplete beam, without providing any solution).
        
\begin{algorithm}[ht]
    \SetAlgoLined
    \DontPrintSemicolon
	\SetKwInOut{Input}{Input}
	\Input{root node}
	\BlankLine
	$D \gets $ 1\;
	\While{stopping criterion not met}{
		Candidates $\gets$ \{root\} \;
        \While{Candidates $\neq \emptyset$}{
            nextLevel $\gets \bigcup_{n \in \text{Candidates}} \text{children}(n)$\;
            Candidates $\gets$ best $D$ nodes among nextLevel\;
        }
        $D \gets D \times $ 2\;
	}
\caption{Iterative Beam Search algorithm \label{alg:BS}}
\end{algorithm}

\section{Numerical results}\label{sec:num}

In this section, we perform various experiments to evaluate the efficiency of the algorithms discussed in the previous sections. In Subsection \ref{ss:num_makespan}, we present numerical results obtained in the makespan minimization version and Subsection \ref{ss:num_flowtime}, the results obtained in the flowtime minimization version. All algorithms have been implemented in C++ and executed on an Intel(R) Core(TM) i5-3470 CPU @3.20GHz with 8GB RAM. As the CPU has 4 physical cores, we ran 4 tests in parallel to obtain results faster. For both objectives, we study the ARPD (Average Relative Percentage Deviation), defined as follows:
\[ ARPD_{Ia} = \sum_{i \in I} \frac{M_{ai}-M^*_i}{M^*_i} \;.\; \frac{100}{|I|} \]
where $I$ is a set of instances with similar characteristics, $M_{ai}$ corresponds to the objective obtained by algorithm $a$ on instance $i$. And $M^*_i$ the best-so-far solution objective for instance $i$. The ARPD describes the performance of a given algorithm on a given instance type. A positive ARPD implies that the algorithm finds in average, solutions dominated by the best-known ones and a negative ARPD implies that the algorithm can improve on the best-known solutions.

\subsection{Makespan minimization}\label{ss:num_makespan}

We ran each algorithm for $m.n.45$ milliseconds where $n$ is the number of jobs and $m$ the number of machines as it is usually done in the literature for the makespan minimization. We evaluate our algorithms on the famous VFR set of instances \cite{vallada2015new}. That consists of sets of 10 instances with a job number $n \in \{100, 200 \dots 800\}$, a machine number $m \in \{20, 40, 60\}$. The benchmark is commonly used in the literature to evaluate the performance of optimization algorithms. Most instances with 20 machines are closed \cite{gmys2020computationally}. The best-so-far results are an aggregation of the results found in the literature \cite{fernandez2019best,kizilay_variable_2019,gmys2020computationally}. Figure \ref{tab:arpdmakespan} presents the ARPD obtained on the VFR instances and makespan minimization.

\begin{figure}[pos=ht]
    \centering
    \begin{tabular}{l|llll|llll}
            & F $g_1$	& F $g_2$	& F $g_3$	& F $g_4$	& FB $g_1$	& FB $g_2$	& FB $g_3$	& FB $g_4$	\\
        \hline
        VFR100\_20	& 18.70	& 3.45	& 3.79	& 3.43	& 2.29	& 11.05	& 9.84	& 0.58	\\
        VFR100\_40	& 17.41	& 4.62	& 4.62	& 3.68	& 5.83	& 9.67	& 9.39	& 2.25	\\
        VFR100\_60	& 16.54	& 5.18	& 5.07	& 4.79	& 6.35	& 9.50	& 9.52	& 3.68	\\
        VFR200\_20	& 21.38	& 2.93	& 3.59	& 1.78	& 1.30	& 10.59	& 8.50	& 1.75	\\
        VFR200\_40	& 21.93	& 4.69	& 4.82	& 2.56	& 5.73	& 14.68	& 14.50	& 0.24	\\
        VFR200\_60	& 20.27	& 5.35	& 5.51	& 2.77	& 8.13	& 15.75	& 14.56	& 1.86	\\
        VFR300\_20	& 22.44	& 2.02	& 2.72	& 1.72	& 0.83	& 7.40	& 5.53	& 1.43	\\
        VFR300\_40	& 22.41	& 4.41	& 4.73	& 1.09	& 5.60	& 16.79	& 14.29	& -0.80	\\
        VFR300\_60	& 20.88	& 5.22	& 5.66	& 1.55	& 8.01	& 12.77	& 12.65	& 0.85	\\
        VFR400\_20	& 22.68	& 1.98	& 2.52	& 1.86	& 0.76	& 4.32	& 3.20	& 1.40	\\
        VFR400\_40	& 24.34	& 4.06	& 4.53	& 0.70	& 5.67	& 16.06	& 14.95	& -0.97	\\
        VFR400\_60	& 21.91	& 5.54	& 6.01	& 0.82	& 7.91	& 17.54	& 16.36	& 0.16	\\
        VFR500\_20	& 24.42	& 1.57	& 2.43	& 1.63	& 0.50	& 3.51	& 3.01	& 1.30	\\
        VFR500\_40	& 24.08	& 3.26	& 3.83	& 0.70	& 4.82	& 14.27	& 12.79	& -0.86	\\
        VFR500\_60	& 22.42	& 5.88	& 6.12	& 0.58	& 8.41	& 16.59	& 15.53	& -0.59	\\
        VFR600\_20	& 23.39	& 1.21	& 2.06	& 1.49	& 0.50	& 3.32	& 2.38	& 1.26	\\
        VFR600\_40	& 24.47	& 3.46	& 3.83	& 0.63	& 5.50	& 16.12	& 14.46	& -0.65	\\
        VFR600\_60	& 22.94	& 5.46	& 5.89	& 0.40	& 7.94	& 13.22	& 12.39	& -0.91	\\
        VFR700\_20	& 24.51	& 1.38	& 2.05	& 1.63	& 0.30	& 2.49	& 1.98	& 1.20	\\
        VFR700\_40	& 25.27	& 3.19	& 3.91	& 0.48	& 4.47	& 13.73	& 13.14	& -0.27	\\
        VFR700\_60	& 23.11	& 5.55	& 5.82	& 0.07	& 8.06	& 15.92	& 14.20	& -1.19	\\
        VFR800\_20	& 24.51	& 1.14	& 1.93	& 1.43	& 0.16	& 2.05	& 1.81	& 1.09	\\
        VFR800\_40	& 25.12	& 3.30	& 4.08	& 0.32	& 4.18	& 13.95	& 13.42	& 0.09	\\
        VFR800\_60	& 23.89	& 5.90	& 6.40	& 0.05	& 7.89	& 15.91	& 14.35	& -1.25	\\
        \hline
        nb new-best-known & 0    & 0     & 0     & 20    & 0     & 0     & 0     & 104   \\
    \end{tabular}
    \caption{Average Relative Percentage Deviation (ARPD) of all the presented algorithms on the VFR instances and the makespan minimization version. The Forward branching is denoted by F (FB for the Forward and Backward (bi-directional) branching). We remind that $g_1$ denotes the bound guide, $g_2$ the idle time, $g_3$ the alpha guide and $g_4$ the weighted alpha guide.}
    \label{tab:arpdmakespan}
\end{figure}

\paragraph{discussions:}
Regarding the forward branching procedures, we observe a significant improvement by including the idle time in the guide and obtain the best results by including a weighted idle time within the guide (similarly to the principles presented in the LR heuristic \cite{liu2001constructive}). Indeed, ARPD ranges from $17\%$ to $25\%$ for the bound guide, and goes down between $1\%$ to $5\%$ for the idle time only and the guide combining the idle time and the bound. Finally, the best results for the forward search are obtained using a weighted idle time and bound guide. Sometimes with an ARPD close to $0\%$ on large instances (meaning it competes with all the best results obtained in the literature), even finding 20 new-best-known solutions. We note that this result is interesting as this algorithm ``only'' combines ideas from the LR heuristic and an iterative beam search. Thus without using components present in recent meta-heuristic state-of-the-art algorithms like local-search moves, the iterated-greedy algorithm, or, NEH-based insertion schemes.

Regarding the bi-directional branching procedures, we observe that the bound guide performs well in most cases, from $0.16\%$ to $8\%$ ARPD. However, using the idle time in the guide (idle time only or idle time combined with the bound) decreases the performance of the algorithm (performances ranging from $2\%$ to $17\%$). It seems to indicate that the idle time is a less efficient guide than the bound for this branching strategy. However, the weighted idle time proves to be a significant bonus and largely improves the quality of the solutions, from $-1.25\%$ to $3\%$ ARPD, finding 100 new-best-known solutions on 160 open instances. Again ``only'' by combining simple ideas (in this case, the LR guidance strategy, the iterative beam search and the bi-directional branching).

\subsection{Flowtime minimization}\label{ss:num_flowtime}

In the literature, algorithms are executed for exactly 1 hour per instance \cite{andrade2019minimizing}, which implies a lot of wasted time on the smallest instances. In this paper, we perform experiments using the following time limit: $m.n.360$ milliseconds where $n$ is the number of jobs and $m$ the number of machines. It allows spending less time on small instances where it is not needed and exactly 1 hour on the largest ones. We show that even using less time, our algorithm can compete with the state-of-the-art and even returns new best-known solutions on most open instances. We perform the comparison using the well-known Taillard dataset \cite{taillard1993benchmarks} that contains instances ranging from 20 to 500 jobs and 5 to 20 machines. Each class of instance contains 10 instances. As the bi-directional branching is not suited for the flowtime minimization (due to the objective structure) and the impact of different variants of the LR heuristic guidance have been already performed in the literature \cite{fernandez2017beam}, we only study the impact of the iterative beam search using LR-inspired guidance strategies (bound biased by the idle time and the bound biased by the weighted idle time). Figure \ref{tab:arpdflowtime} presents the results obtained for the flowtime minimization on the Taillard dataset.

\begin{figure}[pos=ht]
    \centering
    \begin{tabular}{l|ll}
                & F $g_3$	& F $g_4$	\\
        \hline
        TAI\_20\_5	& 0.00	& 0.00	\\
        TAI\_20\_10	& 0.00	& 0.19	\\
        TAI\_20\_20	& 0.00	& 0.57	\\
        TAI\_50\_5	& 0.27	& 0.07	\\
        TAI\_50\_10	& 0.30	& 0.60	\\
        TAI\_50\_20	& 0.14	& 0.93	\\
        TAI\_100\_5	& -0.10	& -0.15	\\
        TAI\_100\_10	& -0.02	& -0.10	\\
        TAI\_100\_20	& -0.18	& 0.62	\\
        TAI\_200\_10	& -0.30	& -0.50	\\
        TAI\_200\_20	& -0.54	& -0.22	\\
        TAI\_500\_20	& -0.32	& -0.45	\\
        \hline
        nb new-best-known & 51 & 44 \\
    \end{tabular}
    \caption{Average Relative Percentage Deviation (ARPD) of all the presented algorithms on the Taillard instances and the flowtime minimization version. The ``F $g_3$'' algotrithm is an iterative beam search guided by the bound guide biased using the idle time and the ``F $g_4$'' being the iterative beam search guided by the bound guide biased using the weighted idle time.}
    \label{tab:arpdflowtime}
\end{figure}

\paragraph{discussions:} We observe that both algorithms perform well for many instances and find new-best-known solutions on approximately 50/100 open instances). By contrast with the makespan minimization, both guidance strategies are comparable in terms of performance (the weighted idle time did not have a significant impact): sometimes $g_3$ performs better than $g_4$ and vice-versa. We may note that the main difference between our results and the beam search algorithms found in the literature \cite{fernandez2017beam} is that we use an iterative beam search that allows performing beam search with larger if the remaining time allows it. This result seems to indicate that the iterative beam search can be of interest to the community as it reports good results compared to other search strategies.

\section{Conclusions \& perspectives}\label{sec:ccl}

In this paper, we present some iterative beam search algorithms applied to the permutation flowshop problem (makespan and flowtime minimization). These algorithms use branching strategies inspired by the LR heuristic (forward branching) and recent branch-and-bound schemes \cite{gmys2020computationally} (bi-directional branching). We compare several guidance strategies (starting from the bound as commonly done in most branch-and-bounds) to more advanced ones (LR inspired guidance). We show that the combination of all of these components obtains state-of-the-art performance. We report 105/160 new-best-so-far solutions for the permutation flowshop (makespan minimization) on the open instances of the VFR benchmark and 55/100 new-best-so-far solutions for the permutation flowshop (flowtime minimization) on the open instances of the Taillard benchmark. These algorithms compare, and sometimes perform better, than the algorithms based on the NEH branching scheme (which is usually considered as ``the most efficient constructive heuristic for the problem'' \cite{fernandez2019best}) and the iterated greedy algorithm (again considered as ``the most efficient approximate algorithm for the problem'' \cite{fernandez2019best}). We believe that the performance of the bi-directional branching combined to the iterative beam search highlighted in this paper could draw the interest of the community for these techniques as they are rather unexplored, although simple and efficient. Studying these techniques leads to a few other questions: We considered the iterative beam search and showed that it is competitive with classical meta-heuristics for the permutation flowshop. However, many other exist. For instance Iterative Memory Bounded A* \cite{fontan2020packingsolver,libralesso2020anytime}, Beam Stack Search \cite{zhou2005beam}, Anytime Column Search \cite{vadlamudi2012anytime}. To the best of our knowledge, they have not been tested yet for the permutation flowshop. In this paper, we studied the makespan and flowtime minimization criteria and achieved competitive results. Many more flowshop variants have been studied. For instance, the blocking flowshop, the distributed permutation flowshop and many others. It could be interesting to assess the performance of the LR-based beam search on these variants.

\bibliographystyle{cas-model2-names}
\bibliography{pfsp_ibs}

\appendix

\section{Notations}

\begin{itemize}
    \item $J$: all the jobs
    \item $M$: all the machines
    \item $n$: job number ($n=|J|$)
    \item $m$: machine number ($m=|M|$)
    \item $F$ (resp. $B$): all the jobs scheduled in the prefix (resp. suffix)
    \item $\cmax_{f,i}$: first availability of machine $i$ in the forward search
    \item $\cmax_{b,i}$: first availability of machine $i$ in the backward search
    \item $R_i$: remaining processing time on machine $i$. $R_i = \sum_{j \in J \setminus \{F \cup B\}} p_{ij}$
    \item $I_{f,i}$: total idle time on machine $i$ in the forward search
    \item $I_{b,i}$: total idle time on machine $i$ in the backward search
    \item $\alpha$: proportion of scheduled jobs. $\alpha = \frac{|F|+|B|}{|J|}$ on bi-directional branching or $\alpha=\frac{|F|}{|J|}$ on forward branching.
    \item $g_1$: guidance function based on the bound (makespan or flowtime)
    \item $g_2$: guidance function based only by the idle time
    \item $g_3$: guidance function based on both the bound and idle time
    \item $g_4$: guidance function based on both the bound and weighted idle time
\end{itemize}

\section{detailed numerical results}

\begin{figure}[pos=ht]
    \centering
    \footnotesize\selectfont
    \begin{tabular}{ll|llll|llll}
            & best-so-far	& F $g_1$	& F $g_2$	& F $g_3$	& F $g_4$	& FB $g_1$	& FB $g_2$	& FB $g_3$	& FB $g_4$	\\
        \hline
        VFR100\_20\_1	& 6.173	& 7.307	& 6.317	& 6.324	& 6.359	& 6.252	& 7.112	& 7.035	& \textbf{6.172}	\\
        VFR100\_20\_2	& 6.267	& 7.471	& 6.421	& 6.534	& 6.586	& 6.386	& 7.535	& 7.347	& 6.306	\\
        VFR100\_20\_3	& 6.221	& 7.384	& 6.384	& 6.394	& 6.398	& 6.319	& 6.787	& 6.650	& 6.231	\\
        VFR100\_20\_4	& 6.227	& 7.447	& 6.455	& 6.455	& 6.399	& 6.306	& 7.483	& 7.235	& 6.268	\\
        VFR100\_20\_5	& 6.264	& 7.410	& 6.362	& 6.386	& 6.609	& 6.468	& 7.254	& 7.098	& 6.330	\\
        VFR100\_20\_6	& 6.285	& 7.399	& 6.457	& 6.490	& 6.468	& 6.521	& 6.843	& 7.027	& 6.333	\\
        VFR100\_20\_7	& 6.401	& 7.594	& 6.669	& 6.683	& 6.558	& 6.593	& 6.656	& 6.650	& 6.409	\\
        VFR100\_20\_8	& 6.074	& 7.230	& 6.230	& 6.214	& 6.219	& 6.213	& 6.467	& 6.380	& 6.083	\\
        VFR100\_20\_9	& 6.328	& 7.470	& 6.719	& 6.747	& 6.464	& 6.472	& 6.542	& 6.500	& 6.426	\\
        VFR100\_20\_10	& 6.125	& 7.314	& 6.502	& 6.504	& 6.441	& 6.269	& 6.572	& 6.575	& 6.167	\\
        VFR100\_40\_1	& 7.846	& 9.303	& 8.233	& 8.214	& 8.230	& 8.300	& 8.558	& 8.640	& 8.010	\\
        VFR100\_40\_2	& 7.976	& 9.355	& 8.375	& 8.377	& 8.121	& 8.443	& 8.945	& 8.888	& 8.091	\\
        VFR100\_40\_3	& 7.894	& 9.366	& 8.258	& 8.255	& 8.142	& 8.331	& 8.846	& 8.867	& 8.018	\\
        VFR100\_40\_4	& 7.913	& 9.254	& 8.171	& 8.156	& 8.379	& 8.336	& 8.268	& 8.240	& 8.154	\\
        VFR100\_40\_5	& 7.997	& 9.275	& 8.472	& 8.475	& 8.353	& 8.481	& 8.522	& 8.476	& 8.108	\\
        VFR100\_40\_6	& 7.993	& 9.350	& 8.339	& 8.345	& 8.288	& 8.432	& 8.374	& 8.332	& 8.279	\\
        VFR100\_40\_7	& 7.980	& 9.356	& 8.317	& 8.335	& 8.206	& 8.523	& 8.298	& 8.284	& 8.344	\\
        VFR100\_40\_8	& 7.957	& 9.263	& 8.281	& 8.288	& 8.244	& 8.413	& 8.926	& 8.947	& 8.122	\\
        VFR100\_40\_9	& 7.888	& 9.332	& 8.208	& 8.208	& 8.174	& 8.361	& 9.142	& 9.052	& 7.987	\\
        VFR100\_40\_10	& 7.917	& 9.322	& 8.371	& 8.373	& 8.141	& 8.372	& 9.150	& 9.080	& 8.036	\\
        VFR100\_60\_1	& 9.353	& 10.808	& 9.946	& 9.935	& 9.868	& 10.050	& 10.285	& 10.287	& 9.757	\\
        VFR100\_60\_2	& 9.567	& 11.151	& 10.095	& 10.023	& 9.968	& 10.081	& 10.160	& 10.155	& 9.738	\\
        VFR100\_60\_3	& 9.349	& 11.033	& 9.823	& 9.817	& 9.743	& 9.902	& 10.681	& 10.690	& 9.656	\\
        VFR100\_60\_4	& 9.403	& 11.069	& 9.919	& 9.949	& 9.911	& 9.981	& 10.368	& 10.387	& 9.613	\\
        VFR100\_60\_5	& 9.431	& 10.917	& 9.916	& 9.929	& 9.740	& 10.039	& 9.976	& 9.912	& 9.966	\\
        VFR100\_60\_6	& 9.630	& 11.279	& 10.044	& 10.044	& 10.088	& 10.234	& 10.572	& 10.517	& 10.016	\\
        VFR100\_60\_7	& 9.346	& 10.933	& 9.852	& 9.874	& 9.861	& 10.008	& 10.344	& 10.414	& 9.654	\\
        VFR100\_60\_8	& 9.523	& 11.189	& 10.041	& 10.046	& 9.967	& 10.158	& 10.344	& 10.387	& 9.843	\\
        VFR100\_60\_9	& 9.488	& 10.866	& 10.008	& 9.954	& 10.050	& 10.082	& 10.326	& 10.391	& 10.070	\\
        VFR100\_60\_10	& 9.572	& 11.073	& 9.918	& 9.889	& 10.001	& 10.134	& 10.588	& 10.526	& 9.829	\\
        VFR200\_20\_1	& 11.272	& 13.546	& 11.436	& 11.553	& 11.513	& 11.473	& 11.605	& 11.549	& 12.213	\\
        VFR200\_20\_2	& 11.240	& 13.473	& 11.869	& 11.852	& 11.489	& 11.381	& 12.821	& 12.647	& 11.444	\\
        VFR200\_20\_3	& 11.294	& 13.716	& 11.514	& 11.556	& 11.433	& 11.445	& 12.422	& 12.306	& 11.328	\\
        VFR200\_20\_4	& 11.188	& 13.629	& 11.434	& 11.442	& 11.368	& 11.307	& 12.475	& 12.539	& 11.265	\\
        VFR200\_20\_5	& 11.143	& 13.504	& 11.442	& 11.573	& 11.494	& 11.340	& 11.903	& 11.733	& 11.242	\\
        VFR200\_20\_6	& 11.310	& 14.030	& 11.552	& 11.798	& 11.457	& 11.456	& 12.579	& 12.179	& 11.428	\\
        VFR200\_20\_7	& 11.365	& 13.840	& 11.532	& 11.590	& 11.584	& 11.523	& 12.631	& 12.260	& 11.437	\\
        VFR200\_20\_8	& 11.128	& 13.501	& 11.566	& 11.636	& 11.311	& 11.148	& 12.345	& 11.981	& 11.279	\\
        VFR200\_20\_9	& 11.091	& 13.449	& 11.636	& 11.720	& 11.240	& 11.365	& 12.830	& 12.592	& 11.285	\\
        VFR200\_20\_10	& 11.294	& 13.655	& 11.628	& 11.636	& 11.432	& 11.346	& 12.601	& 12.080	& 11.366	\\
        VFR200\_40\_1	& 13.124	& 15.833	& 13.856	& 13.786	& 13.442	& 13.731	& 15.301	& 14.977	& 13.139	\\
        VFR200\_40\_2	& 13.049	& 16.029	& 13.704	& 13.578	& 13.350	& 13.734	& 13.634	& 13.840	& 13.131	\\
        VFR200\_40\_3	& 13.222	& 16.167	& 13.750	& 13.758	& 13.523	& 13.835	& 15.590	& 15.677	& 13.228	\\
        VFR200\_40\_4	& 13.163	& 16.108	& 13.816	& 13.771	& 13.548	& 13.945	& 16.003	& 15.662	& 13.232	\\
        VFR200\_40\_5	& 12.974	& 16.077	& 13.569	& 13.741	& 13.273	& 13.744	& 14.560	& 15.661	& 12.997	\\
        VFR200\_40\_6	& 13.061	& 16.094	& 13.833	& 14.005	& 13.313	& 13.861	& 15.208	& 15.167	& 13.134	\\
        VFR200\_40\_7	& 13.220	& 15.921	& 13.719	& 13.733	& 13.507	& 14.070	& 15.188	& 14.883	& 13.296	\\
        VFR200\_40\_8	& 13.132	& 15.903	& 13.556	& 13.520	& 13.616	& 13.812	& 14.786	& 14.929	& \textbf{13.123}	\\
        VFR200\_40\_9	& 13.033	& 15.939	& 13.706	& 13.754	& 13.429	& 13.921	& 15.665	& 15.172	& 13.051	\\
        VFR200\_40\_10	& 13.146	& 15.799	& 13.760	& 13.793	& 13.480	& 13.982	& 14.452	& 14.160	& \textbf{13.114}	\\
        VFR200\_60\_1	& 14.906	& 18.193	& 15.741	& 15.744	& 15.258	& 16.128	& 18.063	& 17.883	& 15.029	\\
        VFR200\_60\_2	& 14.909	& 18.012	& 16.196	& 16.156	& 15.187	& 16.105	& 17.419	& 17.228	& 15.236	\\
        VFR200\_60\_3	& 15.134	& 17.970	& 15.898	& 15.906	& 15.645	& 16.544	& 17.430	& 17.156	& 15.598	\\
        VFR200\_60\_4	& 14.968	& 17.818	& 15.692	& 15.727	& 15.387	& 16.254	& 16.442	& 16.365	& 15.376	\\
        VFR200\_60\_5	& 15.042	& 18.084	& 15.767	& 15.756	& 15.695	& 16.207	& 17.747	& 17.464	& 15.414	\\
        VFR200\_60\_6	& 14.996	& 17.967	& 15.717	& 15.761	& 15.391	& 16.073	& 17.562	& 17.073	& 15.101	\\
        VFR200\_60\_7	& 15.006	& 17.917	& 15.814	& 15.827	& 15.334	& 16.322	& 17.957	& 17.801	& 15.180	\\
        VFR200\_60\_8	& 14.894	& 18.007	& 15.684	& 15.766	& 15.251	& 16.168	& 16.034	& 15.993	& 15.170	\\
        VFR200\_60\_9	& 14.925	& 18.102	& 15.725	& 15.764	& 15.372	& 16.042	& 17.563	& 17.328	& 15.167	\\
        VFR200\_60\_10	& 14.908	& 17.947	& 15.465	& 15.527	& 15.319	& 16.023	& 17.052	& 17.192	& 15.203	\\
        
        \end{tabular}
        
        \label{tab:full_makespan1}
        \caption{Makespan minimization full results: 100 and 200 jobs}
\end{figure}

\begin{figure}[pos=ht]
    \centering
    \footnotesize\selectfont
    \begin{tabular}{ll|llll|llll}
            & best-so-far	& F $g_1$	& F $g_2$	& F $g_3$	& F $g_4$	& FB $g_1$	& FB $g_2$	& FB $g_3$	& FB $g_4$	\\
        \hline
        VFR300\_20\_1	& 16.089	& 19.716	& 16.249	& 16.356	& 16.336	& 16.207	& 17.343	& 16.970	& 16.278	\\
        VFR300\_20\_2	& 16.483	& 20.454	& 16.888	& 16.928	& 16.793	& 16.638	& 17.818	& 17.770	& 16.735	\\
        VFR300\_20\_3	& 16.129	& 19.805	& 16.365	& 16.561	& 16.414	& 16.166	& 17.637	& 16.907	& 16.340	\\
        VFR300\_20\_4	& 16.168	& 19.634	& 16.377	& 16.575	& 16.356	& 16.308	& 17.829	& 17.346	& 16.368	\\
        VFR300\_20\_5	& 16.283	& 19.889	& 16.701	& 16.738	& 16.508	& 16.369	& 17.755	& 17.477	& 16.524	\\
        VFR300\_20\_6	& 16.021	& 20.082	& 16.264	& 16.333	& 16.464	& 16.237	& 17.306	& 16.900	& 16.292	\\
        VFR300\_20\_7	& 16.244	& 19.974	& 16.689	& 16.741	& 16.507	& 16.373	& 16.643	& 16.643	& 16.705	\\
        VFR300\_20\_8	& 16.369	& 19.789	& 16.688	& 16.919	& 16.558	& 16.524	& 17.844	& 17.408	& 16.487	\\
        VFR300\_20\_9	& 16.324	& 19.827	& 16.725	& 16.839	& 16.767	& 16.441	& 17.167	& 16.958	& 16.473	\\
        VFR300\_20\_10	& 16.780	& 20.256	& 17.252	& 17.341	& 16.977	& 16.983	& 17.582	& 17.510	& 17.010	\\
        VFR300\_40\_1	& 18.199	& 22.319	& 19.114	& 19.025	& 18.255	& 19.130	& 21.391	& 20.768	& \textbf{18.059}	\\
        VFR300\_40\_2	& 18.373	& 22.648	& 19.382	& 19.212	& 18.657	& 19.191	& 21.625	& 21.656	& \textbf{18.218}	\\
        VFR300\_40\_3	& 18.348	& 22.609	& 19.088	& 19.322	& 18.536	& 19.562	& 21.062	& 21.465	& \textbf{18.242}	\\
        VFR300\_40\_4	& 18.227	& 22.412	& 18.759	& 18.864	& 18.539	& 19.407	& 21.573	& 20.973	& \textbf{18.095}	\\
        VFR300\_40\_5	& 18.343	& 22.435	& 19.175	& 19.285	& 18.536	& 19.327	& 22.067	& 21.332	& \textbf{18.195}	\\
        VFR300\_40\_6	& 18.340	& 22.392	& 19.225	& 19.300	& 18.428	& 19.553	& 21.679	& 20.951	& \textbf{18.177}	\\
        VFR300\_40\_7	& 18.396	& 22.311	& 19.166	& 19.188	& 18.733	& 19.461	& 21.678	& 21.004	& \textbf{18.202}	\\
        VFR300\_40\_8	& 18.290	& 22.166	& 19.120	& 19.316	& 18.393	& 19.184	& 21.599	& 20.101	& \textbf{18.187}	\\
        VFR300\_40\_9	& 18.261	& 22.488	& 18.991	& 18.964	& 18.530	& 19.171	& 19.291	& 18.902	& \textbf{18.093}	\\
        VFR300\_40\_10	& 18.286	& 22.307	& 19.123	& 19.255	& 18.452	& 19.331	& 21.832	& 22.076	& \textbf{18.132}	\\
        VFR300\_60\_1	& 20.483	& 24.419	& 21.397	& 21.554	& 20.648	& 22.086	& 21.573	& 21.744	& 20.662	\\
        VFR300\_60\_2	& 20.249	& 24.526	& 21.252	& 21.234	& 20.457	& 21.783	& 23.896	& 23.536	& 20.444	\\
        VFR300\_60\_3	& 20.328	& 24.647	& 21.556	& 21.740	& 20.621	& 22.050	& 24.072	& 24.053	& 20.468	\\
        VFR300\_60\_4	& 20.293	& 24.520	& 21.321	& 21.391	& 20.467	& 22.027	& 23.651	& 23.176	& 20.564	\\
        VFR300\_60\_5	& 20.200	& 24.549	& 21.436	& 21.549	& 20.801	& 21.738	& 23.405	& 23.159	& 20.235	\\
        VFR300\_60\_6	& 20.280	& 24.383	& 21.367	& 21.415	& 20.621	& 21.713	& 21.890	& 21.998	& 20.400	\\
        VFR300\_60\_7	& 20.358	& 24.822	& 21.779	& 21.737	& 20.922	& 22.099	& 21.531	& 21.726	& 20.638	\\
        VFR300\_60\_8	& 20.319	& 24.576	& 21.236	& 21.378	& 20.566	& 22.085	& 23.574	& 23.972	& 20.647	\\
        VFR300\_60\_9	& 20.405	& 24.744	& 21.159	& 21.431	& 20.645	& 21.905	& 24.039	& 23.736	& 20.503	\\
        VFR300\_60\_10	& 20.385	& 24.561	& 21.411	& 21.370	& 20.698	& 22.101	& 21.616	& 21.898	& 20.458	\\
        VFR400\_20\_1	& 21.042	& 25.934	& 21.577	& 21.639	& 21.215	& 21.116	& 22.145	& 21.766	& 21.383	\\
        VFR400\_20\_2	& 21.346	& 26.270	& 21.693	& 21.784	& 21.795	& 21.553	& 22.493	& 22.112	& 21.611	\\
        VFR400\_20\_3	& 21.380	& 26.822	& 21.910	& 22.012	& 21.742	& 21.523	& 21.726	& 21.749	& 21.979	\\
        VFR400\_20\_4	& 21.200	& 25.776	& 21.567	& 21.689	& 21.628	& 21.307	& 21.523	& 21.516	& 21.432	\\
        VFR400\_20\_5	& 21.399	& 25.910	& 21.953	& 22.269	& 21.829	& 21.613	& 22.744	& 22.152	& 21.666	\\
        VFR400\_20\_6	& 21.134	& 25.799	& 21.402	& 21.670	& 21.399	& 21.345	& 22.269	& 21.823	& 21.311	\\
        VFR400\_20\_7	& 21.507	& 26.199	& 21.998	& 22.040	& 22.084	& 21.664	& 23.064	& 22.548	& 21.825	\\
        VFR400\_20\_8	& 21.198	& 26.071	& 21.432	& 21.527	& 21.710	& 21.320	& 21.728	& 21.620	& 21.478	\\
        VFR400\_20\_9	& 21.236	& 25.898	& 21.714	& 21.743	& 21.668	& 21.508	& 22.472	& 22.441	& 21.480	\\
        VFR400\_20\_10	& 21.456	& 26.513	& 21.869	& 21.895	& 21.796	& 21.564	& 21.942	& 21.990	& 21.705	\\
        VFR400\_40\_1	& 23.393	& 29.121	& 24.225	& 24.328	& 23.602	& 24.563	& 27.809	& 27.111	& \textbf{23.159}	\\
        VFR400\_40\_2	& 23.380	& 29.227	& 24.260	& 24.347	& 23.467	& 24.886	& 24.384	& 25.004	& \textbf{23.055}	\\
        VFR400\_40\_3	& 23.467	& 28.986	& 24.182	& 24.452	& 23.783	& 24.590	& 26.893	& 26.908	& \textbf{23.258}	\\
        VFR400\_40\_4	& 23.269	& 29.285	& 24.277	& 24.365	& \textbf{23.226}	& 25.029	& 28.186	& 27.259	& \textbf{22.896}	\\
        VFR400\_40\_5	& 23.213	& 28.818	& 24.188	& 24.199	& 23.330	& 24.818	& 27.737	& 27.594	& \textbf{22.984}	\\
        VFR400\_40\_6	& 23.298	& 28.837	& 24.212	& 24.376	& 23.400	& 24.300	& 27.429	& 27.585	& \textbf{23.103}	\\
        VFR400\_40\_7	& 23.415	& 29.280	& 24.236	& 24.321	& 23.529	& 24.782	& 28.610	& 27.597	& \textbf{23.197}	\\
        VFR400\_40\_8	& 23.290	& 28.982	& 24.561	& 24.700	& 23.438	& 24.667	& 24.853	& 24.991	& \textbf{23.149}	\\
        VFR400\_40\_9	& 23.424	& 29.172	& 24.537	& 24.608	& 23.764	& 24.507	& 28.395	& 27.655	& \textbf{23.322}	\\
        VFR400\_40\_10	& 23.606	& 28.946	& 24.563	& 24.636	& 23.848	& 24.867	& 26.990	& 26.992	& \textbf{23.362}	\\
        VFR400\_60\_1	& 25.395	& 30.869	& 27.068	& 27.209	& 25.563	& 27.677	& 29.938	& 29.502	& \textbf{25.359}	\\
        VFR400\_60\_2	& 25.549	& 31.091	& 27.035	& 27.034	& 25.737	& 27.770	& 30.060	& 29.120	& 25.636	\\
        VFR400\_60\_3	& 25.707	& 31.170	& 27.079	& 27.467	& 25.793	& 27.647	& 30.053	& 29.892	& \textbf{25.658}	\\
        VFR400\_60\_4	& 25.638	& 30.985	& 27.317	& 27.339	& 25.983	& 27.314	& 29.051	& 29.451	& 25.797	\\
        VFR400\_60\_5	& 25.669	& 31.179	& 26.822	& 26.942	& 26.025	& 27.394	& 30.367	& 29.596	& 25.788	\\
        VFR400\_60\_6	& 25.407	& 30.940	& 26.444	& 26.615	& 25.689	& 27.306	& 30.083	& 29.858	& 25.473	\\
        VFR400\_60\_7	& 25.415	& 31.320	& 26.987	& 26.990	& 25.525	& 27.335	& 30.479	& 30.247	& 25.434	\\
        VFR400\_60\_8	& 25.603	& 31.200	& 27.006	& 27.187	& 25.702	& 27.794	& 30.167	& 30.038	& \textbf{25.509}	\\
        VFR400\_60\_9	& 25.673	& 31.645	& 26.830	& 27.050	& 25.825	& 28.023	& 30.021	& 29.551	& 25.731	\\
        VFR400\_60\_10	& 25.658	& 31.353	& 27.284	& 27.262	& 25.957	& 27.691	& 30.348	& 30.292	& 25.747	\\
        
        \end{tabular}
        
        \label{tab:full_makespan2}
        \caption{Makespan minimization full results: 300 and 400 jobs}
\end{figure}

\begin{figure}[pos=ht]
    \centering
    \footnotesize\selectfont
    \begin{tabular}{ll|llll|llll}
            & best-so-far	& F $g_1$	& F $g_2$	& F $g_3$	& F $g_4$	& FB $g_1$	& FB $g_2$	& FB $g_3$	& FB $g_4$	\\
        \hline
        VFR500\_20\_1	& 26.253	& 32.562	& 26.600	& 26.812	& 26.703	& 26.318	& 27.582	& 27.139	& 26.656	\\
        VFR500\_20\_2	& 26.555	& 32.773	& 26.875	& 27.161	& 27.035	& 26.786	& 27.433	& 27.202	& 26.862	\\
        VFR500\_20\_3	& 26.268	& 32.757	& 26.647	& 26.835	& 26.685	& 26.444	& 26.907	& 27.006	& 26.593	\\
        VFR500\_20\_4	& 25.994	& 32.623	& 26.551	& 26.949	& 26.488	& 26.102	& 26.744	& 26.715	& 26.295	\\
        VFR500\_20\_5	& 26.703	& 33.177	& 26.915	& 27.198	& 27.048	& 26.843	& 27.271	& 27.187	& 27.169	\\
        VFR500\_20\_6	& 26.325	& 32.944	& 27.027	& 27.247	& 26.854	& 26.374	& 27.506	& 27.314	& 26.544	\\
        VFR500\_20\_7	& 26.313	& 32.861	& 26.693	& 26.863	& 26.749	& 26.474	& 27.105	& 27.208	& 26.723	\\
        VFR500\_20\_8	& 26.217	& 32.349	& 26.570	& 26.703	& 26.555	& 26.420	& 26.803	& 26.803	& 26.559	\\
        VFR500\_20\_9	& 26.345	& 33.100	& 26.810	& 26.873	& 26.777	& 26.466	& 27.473	& 26.759	& 26.719	\\
        VFR500\_20\_10	& 26.034	& 32.078	& 26.435	& 26.748	& 26.410	& 26.101	& 27.414	& 27.588	& 26.302	\\
        VFR500\_40\_1	& 28.402	& 35.162	& 29.766	& 29.764	& \textbf{28.331}	& 30.083	& 29.277	& 29.905	& \textbf{28.183}	\\
        VFR500\_40\_2	& 28.613	& 35.761	& 29.421	& 29.714	& 29.011	& 29.631	& 33.786	& 32.957	& \textbf{28.310}	\\
        VFR500\_40\_3	& 28.526	& 36.006	& 29.575	& 29.713	& 28.653	& 29.672	& 33.521	& 32.471	& \textbf{28.401}	\\
        VFR500\_40\_4	& 28.615	& 34.798	& 29.289	& 29.361	& 28.803	& 30.232	& 33.067	& 32.001	& \textbf{28.378}	\\
        VFR500\_40\_5	& 28.579	& 35.521	& 29.516	& 29.735	& 28.844	& 29.753	& 33.390	& 33.081	& \textbf{28.289}	\\
        VFR500\_40\_6	& 28.432	& 34.683	& 29.435	& 29.633	& 28.643	& 29.568	& 33.687	& 33.024	& \textbf{28.138}	\\
        VFR500\_40\_7	& 28.553	& 35.589	& 29.683	& 29.853	& 28.704	& 30.068	& 30.463	& 30.460	& \textbf{28.307}	\\
        VFR500\_40\_8	& 28.488	& 35.555	& 29.337	& 29.643	& 28.844	& 30.000	& 32.417	& 32.260	& \textbf{28.299}	\\
        VFR500\_40\_9	& 28.640	& 35.728	& 29.467	& 29.448	& 28.839	& 30.413	& 32.792	& 32.856	& \textbf{28.407}	\\
        VFR500\_40\_10	& 28.644	& 35.448	& 29.294	& 29.562	& 28.813	& 29.820	& 33.841	& 33.006	& \textbf{28.318}	\\
        VFR500\_60\_1	& 30.682	& 38.110	& 32.619	& 32.681	& 30.848	& 33.467	& 36.212	& 35.569	& \textbf{30.491}	\\
        VFR500\_60\_2	& 30.664	& 37.489	& 32.537	& 32.344	& 30.924	& 33.207	& 35.416	& 34.622	& \textbf{30.532}	\\
        VFR500\_60\_3	& 30.852	& 37.768	& 33.114	& 33.011	& 30.979	& 33.486	& 36.670	& 35.404	& \textbf{30.671}	\\
        VFR500\_60\_4	& 30.793	& 37.687	& 32.518	& 32.819	& 31.145	& 33.130	& 35.390	& 35.383	& \textbf{30.672}	\\
        VFR500\_60\_5	& 30.763	& 37.624	& 32.553	& 32.475	& 31.051	& 33.527	& 36.672	& 36.371	& \textbf{30.540}	\\
        VFR500\_60\_6	& 30.788	& 37.843	& 32.524	& 32.614	& 31.069	& 33.482	& 35.603	& 35.779	& \textbf{30.597}	\\
        VFR500\_60\_7	& 30.826	& 37.800	& 32.687	& 32.932	& 30.893	& 33.532	& 35.810	& 36.018	& \textbf{30.528}	\\
        VFR500\_60\_8	& 30.837	& 37.261	& 32.696	& 32.807	& \textbf{30.808}	& 33.202	& 36.844	& 36.117	& \textbf{30.584}	\\
        VFR500\_60\_9	& 30.805	& 37.763	& 32.424	& 32.589	& 30.836	& 33.702	& 35.728	& 35.279	& \textbf{30.645}	\\
        VFR500\_60\_10	& 30.866	& 37.553	& 32.316	& 32.452	& 31.119	& 33.030	& 34.603	& 35.138	& \textbf{30.787}	\\
        VFR600\_20\_1	& 31.303	& 38.430	& 31.801	& 32.271	& 31.774	& 31.402	& 32.495	& 32.088	& 31.600	\\
        VFR600\_20\_2	& 31.281	& 38.305	& 31.675	& 31.953	& 31.887	& 31.731	& 32.781	& 32.394	& 31.753	\\
        VFR600\_20\_3	& 31.374	& 38.812	& 31.817	& 32.104	& 31.694	& 31.501	& 32.521	& 32.066	& 31.670	\\
        VFR600\_20\_4	& 31.417	& 38.879	& 31.648	& 31.718	& 31.702	& 31.669	& 32.000	& 31.857	& 31.759	\\
        VFR600\_20\_5	& 31.354	& 38.842	& 31.590	& 31.831	& 31.799	& 31.450	& 32.287	& 31.983	& 31.769	\\
        VFR600\_20\_6	& 31.613	& 38.684	& 31.974	& 32.243	& 32.112	& 31.792	& 32.094	& 32.058	& 32.035	\\
        VFR600\_20\_7	& 31.461	& 38.728	& 31.995	& 32.188	& 31.962	& 31.530	& 32.957	& 32.432	& 31.951	\\
        VFR600\_20\_8	& 31.414	& 38.590	& 31.825	& 32.097	& 31.991	& 31.528	& 32.468	& 32.292	& 31.701	\\
        VFR600\_20\_9	& 31.473	& 39.149	& 31.638	& 31.881	& 31.960	& 31.511	& 32.611	& 32.168	& 32.052	\\
        VFR600\_20\_10	& 31.021	& 38.671	& 31.528	& 31.890	& 31.508	& 31.156	& 31.910	& 31.828	& 31.361	\\
        VFR600\_40\_1	& 33.683	& 41.668	& 34.398	& 34.702	& 33.991	& 35.621	& 38.040	& 37.455	& \textbf{33.385}	\\
        VFR600\_40\_2	& 33.405	& 41.752	& 34.639	& 34.843	& 33.654	& 35.047	& 39.863	& 39.017	& \textbf{33.237}	\\
        VFR600\_40\_3	& 33.713	& 41.633	& 35.217	& 35.230	& 33.957	& 35.529	& 37.783	& 38.386	& \textbf{33.587}	\\
        VFR600\_40\_4	& 33.584	& 41.596	& 34.554	& 34.510	& \textbf{33.544}	& 35.394	& 37.975	& 37.514	& \textbf{33.254}	\\
        VFR600\_40\_5	& 33.401	& 41.423	& 34.637	& 34.815	& 33.615	& 34.932	& 39.009	& 38.767	& \textbf{33.220}	\\
        VFR600\_40\_6	& 33.626	& 42.297	& 34.312	& 34.520	& 33.869	& 35.259	& 38.919	& 38.553	& \textbf{33.420}	\\
        VFR600\_40\_7	& 33.545	& 42.007	& 35.204	& 35.289	& 33.725	& 35.658	& 40.408	& 38.749	& \textbf{33.413}	\\
        VFR600\_40\_8	& 33.298	& 41.441	& 34.812	& 34.814	& 33.397	& 35.294	& 38.126	& 38.396	& \textbf{33.078}	\\
        VFR600\_40\_9	& 33.567	& 41.588	& 34.623	& 34.849	& 33.839	& 35.529	& 39.483	& 38.086	& \textbf{33.250}	\\
        VFR600\_40\_10	& 33.473	& 41.931	& 34.483	& 34.571	& 33.816	& 35.458	& 39.720	& 38.830	& \textbf{33.284}	\\
        VFR600\_60\_1	& 35.976	& 43.980	& 37.649	& 37.956	& 36.009	& 38.931	& 37.908	& 38.353	& \textbf{35.920}	\\
        VFR600\_60\_2	& 35.923	& 44.098	& 37.462	& 37.834	& \textbf{35.836}	& 38.626	& 42.296	& 40.873	& \textbf{35.561}	\\
        VFR600\_60\_3	& 35.917	& 44.400	& 37.718	& 37.742	& 36.350	& 38.643	& 41.393	& 41.845	& \textbf{35.670}	\\
        VFR600\_60\_4	& 36.000	& 44.670	& 38.001	& 37.889	& 36.227	& 38.731	& 42.377	& 42.419	& \textbf{35.640}	\\
        VFR600\_60\_5	& 36.004	& 44.049	& 38.112	& 38.201	& \textbf{35.902}	& 38.630	& 41.973	& 40.904	& \textbf{35.606}	\\
        VFR600\_60\_6	& 35.943	& 44.676	& 38.074	& 38.258	& 36.042	& 38.814	& 42.807	& 40.751	& \textbf{35.529}	\\
        VFR600\_60\_7	& 35.965	& 43.729	& 37.971	& 38.011	& 36.265	& 38.802	& 38.278	& 39.218	& \textbf{35.717}	\\
        VFR600\_60\_8	& 35.894	& 43.914	& 38.344	& 38.570	& 36.052	& 38.827	& 38.491	& 38.780	& \textbf{35.499}	\\
        VFR600\_60\_9	& 35.987	& 44.365	& 38.297	& 38.627	& 36.328	& 39.316	& 39.006	& 39.172	& \textbf{35.588}	\\
        VFR600\_60\_10	& 35.943	& 44.166	& 37.551	& 37.658	& 35.974	& 38.765	& 42.554	& 41.788	& \textbf{35.563}	\\
        
        \end{tabular}
        
        \label{tab:full_makespan3}
        \caption{Makespan minimization full results: 500 and 600 jobs}
\end{figure}

\begin{figure}[pos=ht]
    \centering
    \footnotesize\selectfont
    \begin{tabular}{ll|llll|llll}
            & best-so-far	& F $g_1$	& F $g_2$	& F $g_3$	& F $g_4$	& FB $g_1$	& FB $g_2$	& FB $g_3$	& FB $g_4$	\\
        \hline
        VFR700\_20\_1	& 36.285	& 45.173	& 36.881	& 37.169	& 36.812	& 36.496	& 37.447	& 37.104	& 36.916	\\
        VFR700\_20\_2	& 36.220	& 44.898	& 36.563	& 36.884	& 36.820	& 36.355	& 36.684	& 37.310	& 36.676	\\
        VFR700\_20\_3	& 36.419	& 44.634	& 37.156	& 37.263	& 37.107	& 36.594	& 37.326	& 37.121	& 36.971	\\
        VFR700\_20\_4	& 36.361	& 45.075	& 36.800	& 36.907	& 36.777	& 36.361	& 37.556	& 37.084	& 36.619	\\
        VFR700\_20\_5	& 36.496	& 46.158	& 37.076	& 37.518	& 37.128	& 36.647	& 37.452	& 37.169	& 36.956	\\
        VFR700\_20\_6	& 36.556	& 45.744	& 37.030	& 37.433	& 37.219	& 36.558	& 37.376	& 37.073	& 36.927	\\
        VFR700\_20\_7	& 36.540	& 45.561	& 36.945	& 37.235	& 37.055	& 36.586	& 37.506	& 37.293	& 36.912	\\
        VFR700\_20\_8	& 36.418	& 44.983	& 36.938	& 37.205	& 37.209	& 36.527	& 37.199	& 37.128	& 36.790	\\
        VFR700\_20\_9	& 36.212	& 45.545	& 36.880	& 37.025	& 36.923	& 36.329	& 37.482	& 37.059	& 36.634	\\
        VFR700\_20\_10	& 36.362	& 45.284	& 36.636	& 36.687	& 36.762	& 36.505	& 36.898	& 36.737	& 36.839	\\
        VFR700\_40\_1	& 38.767	& 48.140	& 39.786	& 39.882	& 39.053	& 40.619	& 43.966	& 43.673	& \textbf{38.573}	\\
        VFR700\_40\_2	& 38.560	& 48.750	& 39.855	& 40.379	& \textbf{38.548}	& 39.797	& 40.433	& 45.233	& \textbf{38.316}	\\
        VFR700\_40\_3	& 38.460	& 48.388	& 39.710	& 39.852	& 38.692	& 40.847	& 45.858	& 44.636	& \textbf{38.261}	\\
        VFR700\_40\_4	& 38.597	& 48.549	& 39.597	& 40.064	& 38.799	& 40.163	& 45.958	& 44.192	& \textbf{38.460}	\\
        VFR700\_40\_5	& 38.490	& 47.881	& 39.485	& 39.771	& 38.846	& 40.108	& 39.145	& 40.199	& \textbf{38.339}	\\
        VFR700\_40\_6	& 38.440	& 48.035	& 40.004	& 40.418	& 38.452	& 39.979	& 46.046	& 44.169	& \textbf{38.352}	\\
        VFR700\_40\_7	& 38.355	& 48.340	& 39.312	& 39.589	& 38.531	& 40.452	& 45.127	& 43.757	& \textbf{38.189}	\\
        VFR700\_40\_8	& 38.817	& 48.138	& 40.213	& 40.298	& 39.106	& 40.580	& 45.685	& 44.259	& \textbf{38.778}	\\
        VFR700\_40\_9	& 38.569	& 48.418	& 39.637	& 39.722	& 38.854	& 40.270	& 40.997	& 41.780	& 38.825	\\
        VFR700\_40\_10	& 38.712	& 48.596	& 40.482	& 40.860	& 38.752	& 40.190	& 45.522	& 44.575	& \textbf{38.635}	\\
        VFR700\_60\_1	& 41.192	& 50.359	& 43.381	& 43.904	& 41.430	& 44.966	& 48.093	& 47.887	& \textbf{40.772}	\\
        VFR700\_60\_2	& 41.002	& 50.651	& 43.732	& 43.890	& 41.350	& 44.532	& 48.688	& 46.729	& \textbf{40.664}	\\
        VFR700\_60\_3	& 41.173	& 50.511	& 43.257	& 43.176	& \textbf{40.981}	& 44.841	& 48.492	& 47.555	& \textbf{40.581}	\\
        VFR700\_60\_4	& 41.120	& 50.625	& 43.033	& 42.841	& \textbf{41.008}	& 43.658	& 48.858	& 47.026	& \textbf{40.491}	\\
        VFR700\_60\_5	& 41.167	& 50.535	& 43.605	& 43.772	& \textbf{41.071}	& 44.793	& 48.514	& 46.562	& \textbf{40.641}	\\
        VFR700\_60\_6	& 41.159	& 50.536	& 43.816	& 43.722	& \textbf{41.082}	& 44.476	& 47.269	& 47.437	& \textbf{40.714}	\\
        VFR700\_60\_7	& 40.734	& 50.379	& 43.191	& 43.170	& 40.737	& 44.130	& 47.259	& 47.224	& \textbf{40.331}	\\
        VFR700\_60\_8	& 41.305	& 50.534	& 43.482	& 43.868	& 41.443	& 44.338	& 47.951	& 46.697	& \textbf{40.830}	\\
        VFR700\_60\_9	& 41.111	& 50.864	& 43.404	& 43.546	& 41.163	& 44.446	& 47.823	& 47.859	& \textbf{40.501}	\\
        VFR700\_60\_10	& 41.186	& 51.162	& 43.049	& 43.207	& \textbf{41.182}	& 44.099	& 43.649	& 44.563	& \textbf{40.730}	\\
        VFR800\_20\_1	& 41.413	& 52.067	& 41.877	& 42.181	& 41.976	& 41.521	& 42.044	& 42.158	& 41.843	\\
        VFR800\_20\_2	& 41.282	& 51.449	& 41.657	& 42.107	& 41.957	& 41.323	& 41.886	& 42.119	& 41.623	\\
        VFR800\_20\_3	& 41.319	& 52.365	& 41.683	& 42.024	& 41.818	& 41.367	& 42.245	& 42.006	& 41.693	\\
        VFR800\_20\_4	& 41.375	& 52.005	& 41.878	& 42.106	& 41.923	& 41.452	& 42.226	& 41.942	& 42.056	\\
        VFR800\_20\_5	& 41.626	& 51.981	& 42.209	& 42.666	& 42.250	& 41.704	& 42.496	& 42.479	& 41.959	\\
        VFR800\_20\_6	& 41.919	& 52.373	& 42.644	& 42.918	& 42.556	& 41.919	& 42.640	& 42.549	& 42.416	\\
        VFR800\_20\_7	& 41.395	& 51.177	& 41.645	& 41.747	& 41.742	& 41.541	& 42.014	& 41.725	& 41.812	\\
        VFR800\_20\_8	& 41.390	& 50.761	& 42.006	& 42.605	& 42.093	& 41.505	& 42.518	& 42.443	& 42.022	\\
        VFR800\_20\_9	& 41.697	& 51.467	& 42.197	& 42.316	& 42.435	& 41.697	& 42.916	& 42.706	& 42.113	\\
        VFR800\_20\_10	& 41.489	& 50.943	& 41.847	& 42.245	& 42.073	& 41.557	& 42.408	& 42.272	& 41.908	\\
        VFR800\_40\_1	& 43.466	& 53.765	& 45.065	& 45.367	& 43.691	& 45.513	& 50.262	& 50.510	& \textbf{43.261}	\\
        VFR800\_40\_2	& 43.575	& 54.404	& 44.440	& 44.650	& 43.818	& 45.174	& 48.957	& 48.298	& \textbf{43.289}	\\
        VFR800\_40\_3	& 43.596	& 54.709	& 44.876	& 45.287	& 43.668	& 44.863	& 52.525	& 49.450	& \textbf{43.313}	\\
        VFR800\_40\_4	& 43.743	& 55.055	& 45.069	& 45.659	& 43.828	& 45.522	& 51.327	& 51.034	& \textbf{43.491}	\\
        VFR800\_40\_5	& 43.794	& 54.471	& 45.621	& 45.977	& \textbf{43.750}	& 45.993	& 47.330	& 47.583	& 46.117	\\
        VFR800\_40\_6	& 43.638	& 54.697	& 44.938	& 44.997	& 43.659	& 45.709	& 48.982	& 50.102	& \textbf{43.370}	\\
        VFR800\_40\_7	& 43.484	& 54.649	& 44.562	& 44.641	& 43.787	& 45.310	& 48.764	& 48.729	& \textbf{43.384}	\\
        VFR800\_40\_8	& 43.666	& 54.693	& 45.720	& 46.263	& 43.834	& 45.532	& 50.455	& 49.794	& \textbf{43.469}	\\
        VFR800\_40\_9	& 43.643	& 54.597	& 44.772	& 44.927	& 44.008	& 45.540	& 47.844	& 49.387	& \textbf{43.477}	\\
        VFR800\_40\_10	& 43.630	& 54.773	& 45.571	& 46.286	& \textbf{43.595}	& 45.294	& 50.638	& 49.907	& \textbf{43.446}	\\
        VFR800\_60\_1	& 46.279	& 57.427	& 48.747	& 48.952	& 46.498	& 49.897	& 54.669	& 54.027	& \textbf{45.680}	\\
        VFR800\_60\_2	& 46.232	& 58.017	& 48.663	& 49.075	& \textbf{46.159}	& 50.324	& 53.338	& 52.429	& \textbf{45.728}	\\
        VFR800\_60\_3	& 46.258	& 57.295	& 49.104	& 49.153	& \textbf{46.243}	& 49.567	& 54.299	& 53.912	& \textbf{45.698}	\\
        VFR800\_60\_4	& 46.261	& 57.072	& 48.430	& 48.699	& \textbf{46.218}	& 49.707	& 53.221	& 52.813	& \textbf{45.696}	\\
        VFR800\_60\_5	& 46.164	& 56.947	& 48.705	& 49.005	& 46.526	& 49.924	& 52.729	& 52.635	& \textbf{45.490}	\\
        VFR800\_60\_6	& 46.288	& 56.301	& 49.747	& 50.097	& 46.350	& 49.451	& 53.784	& 53.185	& \textbf{45.504}	\\
        VFR800\_60\_7	& 46.061	& 57.252	& 49.565	& 49.885	& 46.186	& 49.896	& 53.804	& 52.550	& \textbf{45.600}	\\
        VFR800\_60\_8	& 46.257	& 57.812	& 49.104	& 49.183	& \textbf{46.029}	& 49.979	& 52.997	& 52.442	& \textbf{45.964}	\\
        VFR800\_60\_9	& 46.279	& 57.744	& 49.071	& 49.162	& \textbf{46.171}	& 50.146	& 53.303	& 52.208	& \textbf{45.766}	\\
        VFR800\_60\_10	& 46.211	& 56.850	& 48.420	& 48.665	& \textbf{46.162}	& 49.850	& 53.702	& 52.451	& \textbf{45.383}	\\
        
        \end{tabular}
        
        \label{tab:full_makespan4}
        \caption{Makespan minimization full results: 700 and 800 jobs}
\end{figure}
\begin{figure}[pos=ht]
    \centering
    \begin{tabular}{ll|ll}
        & best-so-far	& F $g_3$	& F $g_4$	\\
    \hline
    TA1 / tai20\_5\_0	& 14.033	& 14.033	& 14.033	\\
    TA2 / tai20\_5\_1	& 15.151	& 15.151	& 15.151	\\
    TA3 / tai20\_5\_2	& 13.301	& 13.301	& 13.301	\\
    TA4 / tai20\_5\_3	& 15.447	& 15.447	& 15.447	\\
    TA5 / tai20\_5\_4	& 13.529	& 13.529	& 13.529	\\
    TA6 / tai20\_5\_5	& 13.123	& 13.123	& 13.123	\\
    TA7 / tai20\_5\_6	& 13.548	& 13.548	& 13.548	\\
    TA8 / tai20\_5\_7	& 13.948	& 13.948	& 13.948	\\
    TA9 / tai20\_5\_8	& 14.295	& 14.295	& 14.295	\\
    TA10 / tai20\_5\_9	& 12.943	& 12.943	& 12.943	\\
    TA11 / tai20\_10\_0	& 20.911	& 20.911	& 20.911	\\
    TA12 / tai20\_10\_1	& 22.440	& 22.440	& 22.440	\\
    TA13 / tai20\_10\_2	& 19.833	& 19.833	& 19.872	\\
    TA14 / tai20\_10\_3	& 18.710	& 18.710	& 18.769	\\
    TA15 / tai20\_10\_4	& 18.641	& 18.641	& 18.641	\\
    TA16 / tai20\_10\_5	& 19.245	& 19.245	& 19.350	\\
    TA17 / tai20\_10\_6	& 18.363	& 18.363	& 18.376	\\
    TA18 / tai20\_10\_7	& 20.241	& 20.241	& 20.268	\\
    TA19 / tai20\_10\_8	& 20.330	& 20.330	& 20.455	\\
    TA20 / tai20\_10\_9	& 21.320	& 21.320	& 21.325	\\
    TA21 / tai20\_20\_0	& 33.623	& 33.623	& 33.623	\\
    TA22 / tai20\_20\_1	& 31.587	& 31.587	& 31.726	\\
    TA23 / tai20\_20\_2	& 33.920	& 33.920	& 34.318	\\
    TA24 / tai20\_20\_3	& 31.661	& 31.661	& 31.661	\\
    TA25 / tai20\_20\_4	& 34.557	& 34.557	& 34.726	\\
    TA26 / tai20\_20\_5	& 32.564	& 32.564	& 32.988	\\
    TA27 / tai20\_20\_6	& 32.922	& 32.922	& 33.160	\\
    TA28 / tai20\_20\_7	& 32.412	& 32.412	& 32.412	\\
    TA29 / tai20\_20\_8	& 33.600	& 33.600	& 33.902	\\
    TA30 / tai20\_20\_9	& 32.262	& 32.262	& 32.474	\\
    TA31 / tai50\_5\_0	& 64.802	& 65.020	& 64.817	\\
    TA32 / tai50\_5\_1	& 68.051	& 68.149	& 68.074	\\
    TA33 / tai50\_5\_2	& 63.162	& 63.247	& 63.162	\\
    TA34 / tai50\_5\_3	& 68.226	& 68.241	& 68.226	\\
    TA35 / tai50\_5\_4	& 69.351	& 69.738	& 69.360	\\
    TA36 / tai50\_5\_5	& 66.841	& 66.852	& 66.841	\\
    TA37 / tai50\_5\_6	& 66.253	& 66.427	& 66.277	\\
    TA38 / tai50\_5\_7	& 64.332	& 64.447	& 64.401	\\
    TA39 / tai50\_5\_8	& 62.981	& 63.566	& 63.203	\\
    TA40 / tai50\_5\_9	& 68.770	& 68.845	& 68.834	\\
    
    \end{tabular}
    \label{tab:full_flowtime1}
    \caption{Flowtime minimization full results: TAI1 to TAI40}
\end{figure}

\begin{figure}[pos=ht]
    \centering
    \begin{tabular}{ll|ll}
        & best-so-far	& F $g_3$	& F $g_4$	\\
    \hline
    TA41 / tai50\_10\_0	& 87.114	& 87.140	& 87.353	\\
    TA42 / tai50\_10\_1	& 82.820	& 82.967	& 83.241	\\
    TA43 / tai50\_10\_2	& 79.931	& 80.094	& 80.106	\\
    TA44 / tai50\_10\_3	& 86.446	& 86.475	& 86.637	\\
    TA45 / tai50\_10\_4	& 86.377	& 86.567	& 86.628	\\
    TA46 / tai50\_10\_5	& 86.587	& 86.729	& 87.010	\\
    TA47 / tai50\_10\_6	& 88.750	& 89.720	& 89.759	\\
    TA48 / tai50\_10\_7	& 86.727	& 86.730	& 87.719	\\
    TA49 / tai50\_10\_8	& 85.441	& 85.952	& 86.197	\\
    TA50 / tai50\_10\_9	& 87.998	& 88.392	& 88.722	\\
    TA51 / tai50\_20\_0	& 125.831	& 125.831	& 126.245	\\
    TA52 / tai50\_20\_1	& 119.247	& 119.397	& 120.594	\\
    TA53 / tai50\_20\_2	& 116.459	& 116.536	& 117.727	\\
    TA54 / tai50\_20\_3	& 120.261	& 120.811	& 121.359	\\
    TA55 / tai50\_20\_4	& 118.184	& 118.379	& 119.434	\\
    TA56 / tai50\_20\_5	& 120.586	& 120.637	& 121.526	\\
    TA57 / tai50\_20\_6	& 122.880	& 123.120	& 124.274	\\
    TA58 / tai50\_20\_7	& 122.489	& 122.583	& 123.447	\\
    TA59 / tai50\_20\_8	& 121.872	& 121.872	& 123.022	\\
    TA60 / tai50\_20\_9	& 123.954	& 124.275	& 125.425	\\
    TA61 / tai100\_5\_0	& 253.232	& \textbf{252.821}	& \textbf{252.624}	\\
    TA62 / tai100\_5\_1	& 242.093	& \textbf{241.593}	& \textbf{241.737}	\\
    TA63 / tai100\_5\_2	& 237.832	& \textbf{237.240}	& \textbf{237.345}	\\
    TA64 / tai100\_5\_3	& 227.738	& \textbf{227.420}	& \textbf{227.329}	\\
    TA65 / tai100\_5\_4	& 240.301	& \textbf{240.114}	& \textbf{240.024}	\\
    TA66 / tai100\_5\_5	& 232.342	& \textbf{232.131}	& \textbf{232.008}	\\
    TA67 / tai100\_5\_6	& 240.366	& 240.745	& \textbf{239.843}	\\
    TA68 / tai100\_5\_7	& 230.945	& \textbf{230.304}	& \textbf{230.371}	\\
    TA69 / tai100\_5\_8	& 247.677	& \textbf{247.472}	& \textbf{247.437}	\\
    TA70 / tai100\_5\_9	& 242.933	& 243.254	& 243.062	\\
    TA71 / tai100\_10\_0	& 298.385	& \textbf{298.002}	& \textbf{297.749}	\\
    TA72 / tai100\_10\_1	& 273.826	& 273.852	& \textbf{273.765}	\\
    TA73 / tai100\_10\_2	& 288.114	& 288.275	& \textbf{287.614}	\\
    TA74 / tai100\_10\_3	& 301.044	& \textbf{300.545}	& \textbf{300.601}	\\
    TA75 / tai100\_10\_4	& 284.279	& \textbf{283.961}	& \textbf{283.637}	\\
    TA76 / tai100\_10\_5	& 269.686	& \textbf{269.436}	& \textbf{269.453}	\\
    TA77 / tai100\_10\_6	& 279.463	& 280.681	& 280.467	\\
    TA78 / tai100\_10\_7	& 290.908	& \textbf{290.219}	& \textbf{289.947}	\\
    TA79 / tai100\_10\_8	& 301.970	& \textbf{301.843}	& 302.110	\\
    TA80 / tai100\_10\_9	& 291.283	& 291.439	& \textbf{290.735}	\\
    
    \end{tabular}
    \label{tab:full_flowtime2}
    \caption{Flowtime minimization full results: TAI41 to TAI80}
\end{figure}

\begin{figure}[pos=ht]
    \centering
    \begin{tabular}{ll|ll}
            & best-so-far	& F $g_3$	& F $g_4$	\\
        \hline
        TA81 / tai100\_20\_0	& 365.463	& 366.544	& 368.283	\\
        TA82 / tai100\_20\_1	& 372.449	& \textbf{371.544}	& 374.585	\\
        TA83 / tai100\_20\_2	& 370.027	& \textbf{369.571}	& 373.153	\\
        TA84 / tai100\_20\_3	& 372.393	& \textbf{370.754}	& 374.532	\\
        TA85 / tai100\_20\_4	& 368.915	& \textbf{366.924}	& 370.549	\\
        TA86 / tai100\_20\_5	& 370.908	& 370.950	& 374.142	\\
        TA87 / tai100\_20\_6	& 373.408	& \textbf{372.225}	& 374.978	\\
        TA88 / tai100\_20\_7	& 384.525	& \textbf{383.271}	& 386.348	\\
        TA89 / tai100\_20\_8	& 374.423	& \textbf{374.413}	& 376.850	\\
        TA90 / tai100\_20\_9	& 379.296	& \textbf{378.948}	& 381.513	\\
        TA91 / tai200\_10\_0	& 1.042.494	& \textbf{1.040.290}	& \textbf{1.035.022}	\\
        TA92 / tai200\_10\_1	& 1.028.957	& \textbf{1.025.209}	& \textbf{1.024.879}	\\
        TA93 / tai200\_10\_2	& 1.043.467	& \textbf{1.041.260}	& \textbf{1.037.699}	\\
        TA94 / tai200\_10\_3	& 1.029.244	& \textbf{1.021.739}	& \textbf{1.018.655}	\\
        TA95 / tai200\_10\_4	& 1.029.384	& \textbf{1.027.978}	& \textbf{1.024.342}	\\
        TA96 / tai200\_10\_5	& 999.241	& \textbf{995.394}	& \textbf{994.499}	\\
        TA97 / tai200\_10\_6	& 1.042.663	& \textbf{1.040.074}	& \textbf{1.038.736}	\\
        TA98 / tai200\_10\_7	& 1.035.981	& \textbf{1.034.159}	& \textbf{1.034.056}	\\
        TA99 / tai200\_10\_8	& 1.015.389	& \textbf{1.013.444}	& \textbf{1.012.533}	\\
        TA100 / tai200\_10\_9	& 1.022.277	& \textbf{1.018.518}	& \textbf{1.017.258}	\\
        TA101 / tai200\_20\_0	& 1.223.860	& \textbf{1.210.533}	& \textbf{1.219.147}	\\
        TA102 / tai200\_20\_1	& 1.234.081	& \textbf{1.230.809}	& \textbf{1.233.361}	\\
        TA103 / tai200\_20\_2	& 1.259.866	& \textbf{1.250.456}	& \textbf{1.253.413}	\\
        TA104 / tai200\_20\_3	& 1.228.060	& \textbf{1.221.033}	& \textbf{1.222.571}	\\
        TA105 / tai200\_20\_4	& 1.219.886	& \textbf{1.209.411}	& \textbf{1.215.093}	\\
        TA106 / tai200\_20\_5	& 1.219.432	& \textbf{1.213.883}	& \textbf{1.217.223}	\\
        TA107 / tai200\_20\_6	& 1.234.366	& \textbf{1.232.351}	& 1.237.431	\\
        TA108 / tai200\_20\_7	& 1.240.627	& \textbf{1.229.895}	& \textbf{1.231.867}	\\
        TA109 / tai200\_20\_8	& 1.220.873	& \textbf{1.216.338}	& 1.221.412	\\
        TA110 / tai200\_20\_9	& 1.235.462	& 1.235.641	& 1.238.120	\\
        TA111 / tai500\_20\_0	& 6.558.547	& \textbf{6.542.681}	& \textbf{6.529.190}	\\
        TA112 / tai500\_20\_1	& 6.679.507	& \textbf{6.659.112}	& \textbf{6.656.697}	\\
        TA113 / tai500\_20\_2	& 6.624.893	& \textbf{6.608.608}	& \textbf{6.596.258}	\\
        TA114 / tai500\_20\_3	& 6.649.855	& \textbf{6.623.800}	& \textbf{6.612.598}	\\
        TA115 / tai500\_20\_4	& 6.590.021	& \textbf{6.578.189}	& \textbf{6.576.523}	\\
        TA116 / tai500\_20\_5	& 6.603.691	& \textbf{6.581.804}	& \textbf{6.571.552}	\\
        TA117 / tai500\_20\_6	& 6.576.201	& \textbf{6.551.244}	& \textbf{6.546.293}	\\
        TA118 / tai500\_20\_7	& 6.629.393	& \textbf{6.612.945}	& \textbf{6.605.891}	\\
        TA119 / tai500\_20\_8	& 6.589.205	& \textbf{6.552.881}	& \textbf{6.543.443}	\\
        TA120 / tai500\_20\_9	& 6.626.342	& \textbf{6.606.392}	& \textbf{6.590.453}	\\
        
        \end{tabular}
        \label{tab:full_flowtime3}
        \caption{Flowtime minimization full results: TAI81 to TAI120}
\end{figure}

\end{document}